%% file: arxiv.tex
\documentclass[10pt,twocolumn,letterpaper]{article}

\usepackage[pagenumbers]{cvpr} 

\usepackage{graphicx}
\usepackage{amsmath}
\usepackage{amssymb}
\usepackage{booktabs}

\usepackage{pbox}
\usepackage{calc}
\usepackage{algorithm,algorithmic}
\usepackage{ifthen}
\usepackage{adjustbox}
\usepackage{array,multirow,graphicx}
\usepackage{enumitem}

\usepackage{color, colortbl}
\definecolor{Gray}{gray}{0.9}

\usepackage{listings}
\usepackage[pagebackref,breaklinks,colorlinks]{hyperref}

\usepackage[capitalize]{cleveref}
\crefname{section}{Sec.}{Secs.}
\Crefname{section}{Section}{Sections}
\Crefname{table}{Table}{Tables}
\crefname{table}{Tab.}{Tabs.}

\def\cvprPaperID{725} 
\def\confName{CVPR}
\def\confYear{2023}

\begin{document}

\title{Doubly Right Object Recognition: Visual Reasoning via Prompting}
\title{Doubly Right Object Recognition: \\ Enabling Visual Rationales via ``Why'' Prompting}
\title{Doubly Right Object Recognition: \\ Adapting Foundation Models for Visual Rationales via ``Why'' Prompting}
\title{Doubly Right Object Recognition:  A \emph{Why} Prompt for Visual Rationales}

\author{
Chengzhi Mao$^1$
\hspace{1em}
Revant Teotia$^1$
\hspace{1em}
Amrutha Sundar$^1$
\hspace{1em}
Sachit Menon$^1$ \\
\hspace{1em} Junfeng Yang$^1$
\hspace{1em} Xin Wang$^2$
\hspace{1em} Carl Vondrick$^1$
\\
$^1$Columbia University\hspace{1em} $^2$Microsoft Research\\
{\tt\small \{mcz,rt2891,as6431,sm4934,junfeng,vondrick\}@cs.columbia.edu, wanxin@microsoft.com}}
\maketitle

\begin{abstract} 
Many visual recognition models are evaluated only on their classification accuracy, a metric for which they obtain strong performance. In this paper, we investigate whether computer vision models can also provide correct rationales for their predictions. We propose a ``doubly right'' object recognition benchmark, where the metric requires the model to simultaneously produce both the right labels as well as the right rationales. We find that state-of-the-art visual models, such as CLIP, often provide incorrect rationales for their categorical predictions. However, by transferring the rationales from language models into visual representations through a tailored dataset, we show that we can learn a ``why prompt,'' which adapts large visual representations to produce correct rationales. Visualizations and empirical experiments show that our prompts significantly improve performance on doubly right object recognition, in addition to zero-shot transfer to unseen tasks and datasets.

\end{abstract}

%


\input{src/def}
\input{src/intro}

\input{src/relatedwork}

\input{src/dataset}

\input{src/Experiments}

\input{src/conclusion}

{\small
\bibliographystyle{ieee_fullname}
\bibliography{egbib}
}

\end{document}

%% file: src/def.tex
\def\Blue{\color{blue}}
\def\Purple{\color{purple}}

\def\A{{\bf A}}
\def\a{{\bf a}}
\def\B{{\bf B}}
\def\b{{\bf b}}
\def\C{{\bf C}}
\def\c{{\bf c}}
\def\D{{\bf D}}
\def\d{{\bf d}}
\def\E{{\bf E}}
\def\e{{\bf e}}
\def\f{{\bf f}}
\def\F{{\bf F}}
\def\K{{\bf K}}
\def\k{{\bf k}}
\def\L{{\bf L}}
\def\H{{\bf H}}
\def\h{{\bf h}}
\def\G{{\bf G}}
\def\g{{\bf g}}
\def\I{{\bf I}}
\def\R{{\bf R}}
\def\X{{\bf X}}
\def\Y{{\bf Y}}
\def\OO{{\bf O}}
\def\oo{{\bf o}}
\def\P{{\bf P}}
\def\Q{{\bf Q}}
\def\r{{\bf r}}
\def\s{{\bf s}}
\def\S{{\bf S}}
\def\t{{\bf t}}
\def\T{{\bf T}}
\def\x{{\bf x}}
\def\y{{\bf y}}
\def\z{{\bf z}}
\def\Z{{\bf Z}}
\def\M{{\bf M}}
\def\m{{\bf m}}
\def\n{{\bf n}}
\def\U{{\bf U}}
\def\u{{\bf u}}
\def\V{{\bf V}}
\def\v{{\bf v}}
\def\W{{\bf W}}
\def\w{{\bf w}}
\def\0{{\bf 0}}
\def\1{{\bf 1}}
\def\N{{\bf N}}

\def\AM{{\mathcal A}}
\def\EM{{\mathcal E}}
\def\FM{{\mathcal F}}
\def\TM{{\mathcal T}}
\def\UM{{\mathcal U}}
\def\XM{{\mathcal X}}
\def\YM{{\mathcal Y}}
\def\NM{{\mathcal N}}
\def\OM{{\mathcal O}}
\def\IM{{\mathcal I}}
\def\GM{{\mathcal G}}
\def\PM{{\mathcal P}}
\def\LM{{\mathcal L}}
\def\MM{{\mathcal M}}
\def\DM{{\mathcal D}}
\def\SM{{\mathcal S}}
\def\RB{{\mathbb R}}
\def\EB{{\mathbb E}}

\def\tx{\tilde{\bf x}}
\def\ty{\tilde{\bf y}}
\def\tz{\tilde{\bf z}}
\def\hd{\hat{d}}
\def\HD{\hat{\bf D}}
\def\hx{\hat{\bf x}}
\def\hR{\hat{R}}

\def\Ome{\mbox{\boldmath$\omega$\unboldmath}}
\def\bet{\mbox{\boldmath$\beta$\unboldmath}}
\def\et{\mbox{\boldmath$\eta$\unboldmath}}
\def\ep{\mbox{\boldmath$\epsilon$\unboldmath}}
\def\ph{\mbox{\boldmath$\phi$\unboldmath}}
\def\Pii{\mbox{\boldmath$\Pi$\unboldmath}}
\def\pii{\mbox{\boldmath$\pi$\unboldmath}}
\def\Ph{\mbox{\boldmath$\Phi$\unboldmath}}
\def\Ps{\mbox{\boldmath$\Psi$\unboldmath}}
\def\pss{\mbox{\boldmath$\psi$\unboldmath}}
\def\tha{\mbox{\boldmath$\theta$\unboldmath}}
\def\Tha{\mbox{\boldmath$\Theta$\unboldmath}}
\def\muu{\mbox{\boldmath$\mu$\unboldmath}}
\def\Si{\mbox{\boldmath$\Sigma$\unboldmath}}
\def\Gam{\mbox{\boldmath$\Gamma$\unboldmath}}
\def\gamm{\mbox{\boldmath$\gamma$\unboldmath}}
\def\Lam{\mbox{\boldmath$\Lambda$\unboldmath}}
\def\De{\mbox{\boldmath$\Delta$\unboldmath}}
\def\vps{\mbox{\boldmath$\varepsilon$\unboldmath}}
\def\Up{\mbox{\boldmath$\Upsilon$\unboldmath}}
\def\Lap{\mbox{\boldmath$\LM$\unboldmath}}
\newcommand{\ti}[1]{\tilde{#1}}

\def\tr{\mathrm{tr}}
\def\etr{\mathrm{etr}}
\def\etal{{\em et al.\/}\,}
\newcommand{\indep}{{\;\bot\!\!\!\!\!\!\bot\;}}
\def\argmax{\mathop{\rm argmax}}
\def\argmin{\mathop{\rm argmin}}
\def\vec{\text{vec}}
\def\cov{\text{cov}}
\def\dg{\text{diag}}

\newcommand{\tabref}[1]{Table~\ref{#1}}
\newcommand{\lemref}[1]{Lemma~\ref{#1}}
\newcommand{\thmref}[1]{Theorem~\ref{#1}}
\newcommand{\clmref}[1]{Claim~\ref{#1}}
\newcommand{\crlref}[1]{Corollary~\ref{#1}}
\newcommand{\eqnref}[1]{Eqn.~\ref{#1}}

\newtheorem{remark}{Remark}
\newtheorem{theorem}{Theorem}
\newtheorem{lemma}{Lemma}
\newtheorem{definition}{Definition}

\newtheorem{proposition}{Proposition}

%% file: src/intro.tex
\section{Introduction}


Computer vision models today are able to achieve high accuracy -- sometimes super-human -- at correctly recognizing objects in images. However, most models today are not evaluated on whether they get the prediction right for the right reasons~\cite{he2016deep, VGG_ICLR_15, dosovitskiy2020vit, efficientnet}. Learning models that can explain their own decision is important for building trustworthy systems, especially in applications that require human-machine interactions~\cite{singh2020explainable, ancona2018towards, cnn_mri_attribution_2019, cnn_braintumor_2018}. Rationales that justify the prediction can largely improve user trust~\cite{teach1981analysis}, which is a crucial metric that the visual recognition field should push forward in the future. 

Existing methods in interpretability have investigated how to understand which features contribute to the models' prediction~\cite{Anh_nips2016, olah2017feature_visualization, Simonyan14a_saliency_maps, Shrikumar_icml2017, DeconvNet_eccv14, Smilkov_smoothgrad_17, gradcam_2016}. However, saliency explanations are often imprecise, require domain expertise to understand, and also cannot be evaluated.  \cite{hendricks2016generating, kim2018textual} have instead explored verbal rationales to justify the decision-making. However, they require manual collections of the plausible rationales in the first place, which subsequently are limited to small-scale datasets and tasks~\cite{wah2011caltech, awa}.

\begin{figure}[t]
\centering
  \vspace{-5mm}
\includegraphics[width=0.45\textwidth]{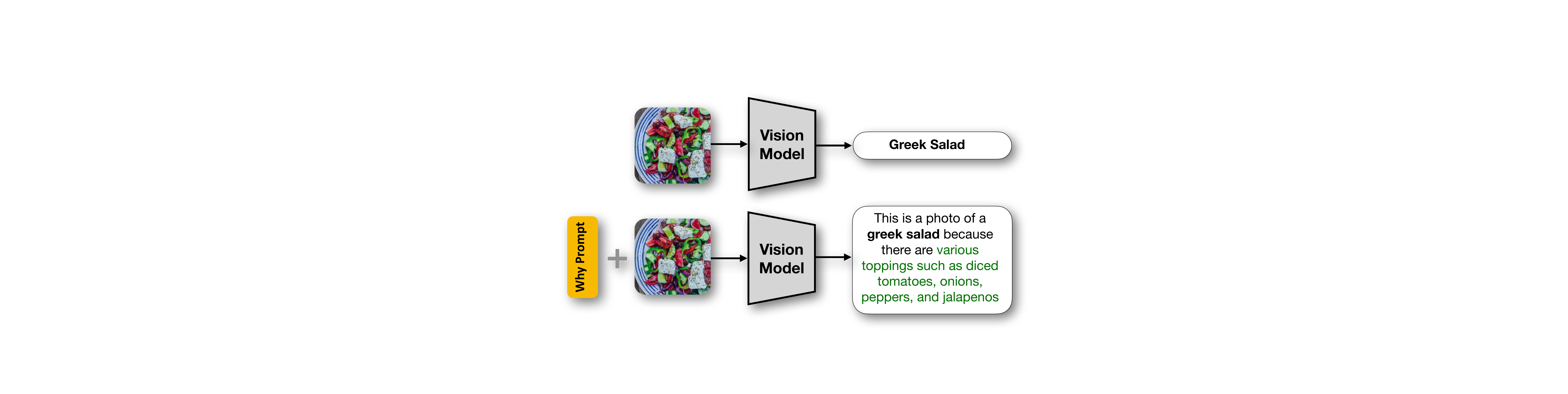}
  \caption{Visual reasoning for doubly right object recognition task. Motivated by prompting in NLP, we learn a \textit{why} prompt from multimodal data, which allows us to instruct visual models to predict both the right category and the correct rationales that justify the prediction.} 
  \label{fig:pca}
  \vspace{-5mm}
\end{figure} 

Scalable methods for explainability have been developed in natural language processing (NLP) through \textit{prompting}. By adding additional instructions to the input, such as the sentence ``think step-by-step,'' language models then output descriptions of their reasoning through the chain of thought process~\cite{wei2022chain}. Since the explanations are verbal, they are easily understandable by people, and since the mechanism emerges without explicit supervision, it is highly scalable. In this paper, we investigate whether visual representations can also explain their reasoning through visual chain-of-thought prompts.

Our paper first introduces a benchmark for doubly right object recognition, where computer vision models must predict both correct categorical labels as well as correct rationales. Our benchmark is large, and covers many categories and datasets. We found that the visual representations do not have double right capability out-of-the-box on our benchmark. The recent large-scale image-language pretrained models~\cite{radford2021learning, singh2022flava} can retrieve open-world language descriptions that are closest to the image embedding in the feature space, serving as verbal explanations.
However, the models often select the wrong rationales.

Instead, we propose a framework to explicitly transfer the chain-of-thought reasoning from NLP models into vision models. We first query the large-scale language model~\cite{GPT3} via the chain-of-thought reasoning for object category, where we obtain language rationales that explain discriminative features for an object. We then collect images containing both the category and the rationale features using Google image search. We then train visual \textit{prompts} to transfer the verbal chain of thought to visual chain of thought with contrastive learning, where features of images and their rationales are pulled together. Our ``why'' prompts obtain up to 26 points gain at doubly right performance when evaluated on our benchmark. In addition, visualizations and quantitative results show that our why prompts zero-shot transfer to unseen tasks and datasets. We believe this ``doubly right'' object recognition task is a future direction which the visual recognition field should go forward for. Our data and code is available at \url{https://github.com/cvlab-columbia/DoubleRight}.

%% file: src/relatedwork.tex
\section{Related Work}

\begin{figure*}[t]
\centering
\includegraphics[width=0.99\textwidth]{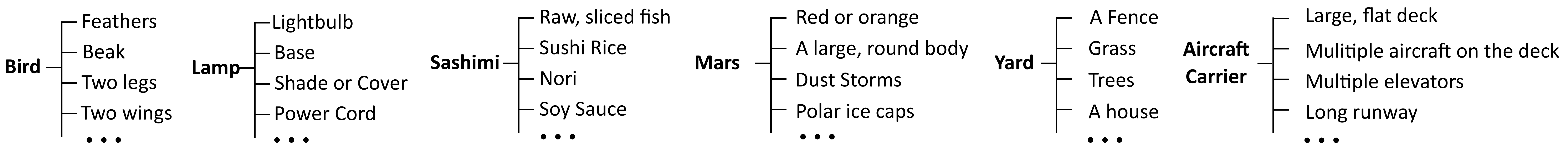}
  \caption{Examples of the rationales generated by prompting GPT3 with the chain of thought reasoning on the features of objects. We show one example from each of the six datasets---CIFAR10, CIFAR100, Food101, Caltech101, SUN, and ImageNet--- we studied. The \textbf{bold} word is the category, and the list follows is the generated rationales. GPT3 can produce reasoning for why an object is as predicted.} 
  \label{fig:language_att}
  \vspace{-5mm}
\end{figure*} 

\noindent\textbf{Explainability.} Visual recognition models achieve high performance on the classification tasks, yet they often provide vague and unreliable interpretations and rationales~\cite{biran2017explanation}. 
There are two lines of research for interpreting image classification using neural networks: feature visualization and language explanation. Feature visualization methods~\cite{Anh_cvpr2017, Anh_nips2016, olah2017feature_visualization, broden_2017} find the inputs that maximize the outputs of learned features. Gradient-based feature visualization~\cite{SHAP_nips2017, LIME_KDD16, Simonyan14a_saliency_maps, Shrikumar_icml2017, DeconvNet_eccv14, Smilkov_smoothgrad_17, gradcam_2016} highlight the input features/pixels in images that are most important for the network to make decisions. Since the visualizations are often abstract, it is hard for a non-expert to understand. In addition, saliency map~\cite{gradcam_2016} highlights regions that may contain overlapping concepts such as color, texture, and shape, which is hard to disentangle. The second line of research uses language-based explanation methods~\cite{hendricks2016generating, kim2018textual} to generate visual explanations. However, those methods require human annotations, which limits their ability to evaluate on a larger scale.


\noindent\textbf{External Knowledge.} Visual models often learn spurious correlations without external knowledge~\cite{mao2022causal}. External knowledge allows models to learn the right features and obtain better transferability~\cite{shen2022k, liu2019knowledge}. Large-scale pretrained language models, such as GPT-3~\cite{GPT3}, contains knowledge and commonsense learned from the Internet~\cite{petroni2019language}.  \cite{wei2022chain} shows that designing the right prompt, such as the chain of thought, improves the model's ability for language reasoning, which we leverage as an external knowledge source.  Other sources of external knowledge include interactions \cite{alahi2016social, chuang2018learning, vicol2018moviegraphs}, physics \cite{mottaghi2016happens, ye2018interpretable}, etc. \cite{menon2022visual} provides descriptions together with the objective category, which improves recognition performance. However, since they do not annotate rationales in their approach, they cannot measure CLIP's ability to produce the right rationales.

\noindent\textbf{Visual Attributes} Several works have studied visual attributes in images ~\cite{SunAttributes_2012, mit_places, deep_fashion_2016, imaterial_fashion_2019, human_attribute_2016, coco_attributes_2016, VAW_2021_CVPR}.  \cite{AwA2_2017} used the visual attributes for animal classification and \cite{facever_pami2011} used them for face verification. However, some of the attributes are spurious correlated~\cite{singla2022salient} with the prediction task. In this work, we aim to generate visual rationales that produce robust visual attributes like shape and parts, instead of spurious features such as the background.

\noindent\textbf{Visual Reasoning.} \cite{zellers2019recognition} proposed a new cognition benchmark that the model needs to predict both the answer and the rationale to be correct. Visual Question Answering \cite{antol2015vqa} performs visual understanding as a QA task, such as questions about COCO images. However, the questions are not asking for rationales to justify object recognition. \cite{zeng2022socratic} shows that multi-modal vision language models can perform zero-shot image-language tasks, such as image captioning. However, all existing visual reasoning work does not directly evaluate object recognition on the rationales they provided.

\noindent\textbf{Prompting.} 
Prompt tuning is a lightweight adaptation method for language task~\cite{CoOP, CoCoOP, zhou2022prompt}. Recently, the computer vision field has adapted the language and proposed the visual prompts to adapt vision models~\cite{visprompt, jia2022visual, sandler2022fine, visualpromptphillp}. Due to their lightweight, it has been shown effective for continuous learning~\cite{conder2022efficient}. One advantage of the visual prompt is that it does not require model access at test time, which is flexible~\cite{visualpromptphillp}. While existing visual prompt methods focus on improving the recognition task performance, we propose to use this lightweight prompt to improve the models' ability to provide visual rationales.


\begin{table}[t]
  \caption{\label{tab:imgnet_num} {List of datasets that include both the model prediction and rationales. Animals with attributes (AWA) collects discriminative attributes for animals. CUB collects verbal rationales on only birds. BDD-X collects rationales for explaining driving scenarios. VAW and Broaden collect a large number of attributes. However, they are often not the right rationales for explaining why objects are classified, such as color. The * indicates the number of annotated video frames. The existing datasets are often small-scale, limited in domains, and not annotating rationales for the object recognition task. Our framework allows the automatic collection of diversified categories over a large scale. Note that the $^+$ name in our dataset, such as CIFAR-10$^+$, indicates that we collect the same set of categories in CIFAR-10 from google with our pipeline. Our benchmark is large, containing more categories and dataset variants than prior methods.} }
  \centering
  \scriptsize
  \begin{tabular}{lllll}
    \toprule
    Dataset      & Number of     & Number of  & Number of & For Right \\
    Name     & Categories     &   Descriptions & Images & Rationales \\
    \midrule
    AWA~\cite{awa} & 50  &  85 & 30,475 & Yes   \\
    CUB~\cite{wah2011caltech, hendricks2016generating} & 200 & N/A & 11,788 & Yes \\
    BDD-X~\cite{kim2018textual} & 906 & 1,668 & 26,000$^*$ & Yes \\
    VAW~\cite{Pham_2021_CVPR} & N/A & 650 & 72,274 & No \\
    Broaden~\cite{netdissect2017} & 584 & 1,300 & 62,476 & No \\
    \midrule
    Ours CIFAR-10$^+$ & 10 & 63 & 2,201 & Yes \\
    Ours CIFAR-100$^+$ & 100 & 540 & 18,318 & Yes \\
    Ours Food101$^+$ & 101 & 435 & 15,212 & Yes \\
    Ours Caltech101$^+$ & 101 & 516 & 16,849 & Yes \\
    Ours SUN$^+$ & 397 & 2,170 & 75,381 & Yes \\
    Ours ImageNet$^+$ & 1000 & 5,810 & 271,016 & Yes \\ 

    \bottomrule
  \end{tabular}
\end{table}

%% file: src/dataset.tex
\section{Language Rationales to Visual Rationales}
We first introduce how to obtain the language rationales for discriminating an object. We then show how we translate the language descriptions into visual images and construct the dataset. We then propose to construct a ``why prompt'' to adapt the large visual models to produce the right rationales for their predictions.



\subsection{Rationale from Language Model}
A standard way of getting visual rationales for image predictions is through manual annotation. However, manual annotation is expensive, especially when applied to large-scale datasets. Recent advances in large-scale language models, such as the GPT3, demonstrate the ability to provide various commonsense knowledge in language. 
By providing chain-of-thought instructions to the language model, \cite{wei2022chain} shows that language models can perform the task of interest as well as produce explanations. We can use large-scale language models as a tool to collect rationales by asking the right language prompts. In addition, since the rationales are presented in language, it is easy to understand, even for non-experts.

Motivated by the recent work in language prompting~\cite{menon2022visual, GPT3}, we propose to ask the language model: what are the valid rationales for an object prediction? Specifically, to obtain rationales for visual objects, we ask GPT3 the following question:
\begin{lstlisting}[breakatwhitespace=true]
Q: What are useful visual features for distinguishing a {category name} in a photo?
A: There are several useful visual features to tell there is a {category name} in a photo:
-
\end{lstlisting}
where GPT3 automatically generates the answer for us, which scales to large, unseen categories. In Figure~\ref{fig:language_att}, we show random examples of the rationales generated by querying GPT3. Each category is provided with a list of discriminative features for the category, which is consistent with how people explain their predictions.

\begin{figure}[t]
\centering
\includegraphics[width=0.45\textwidth]{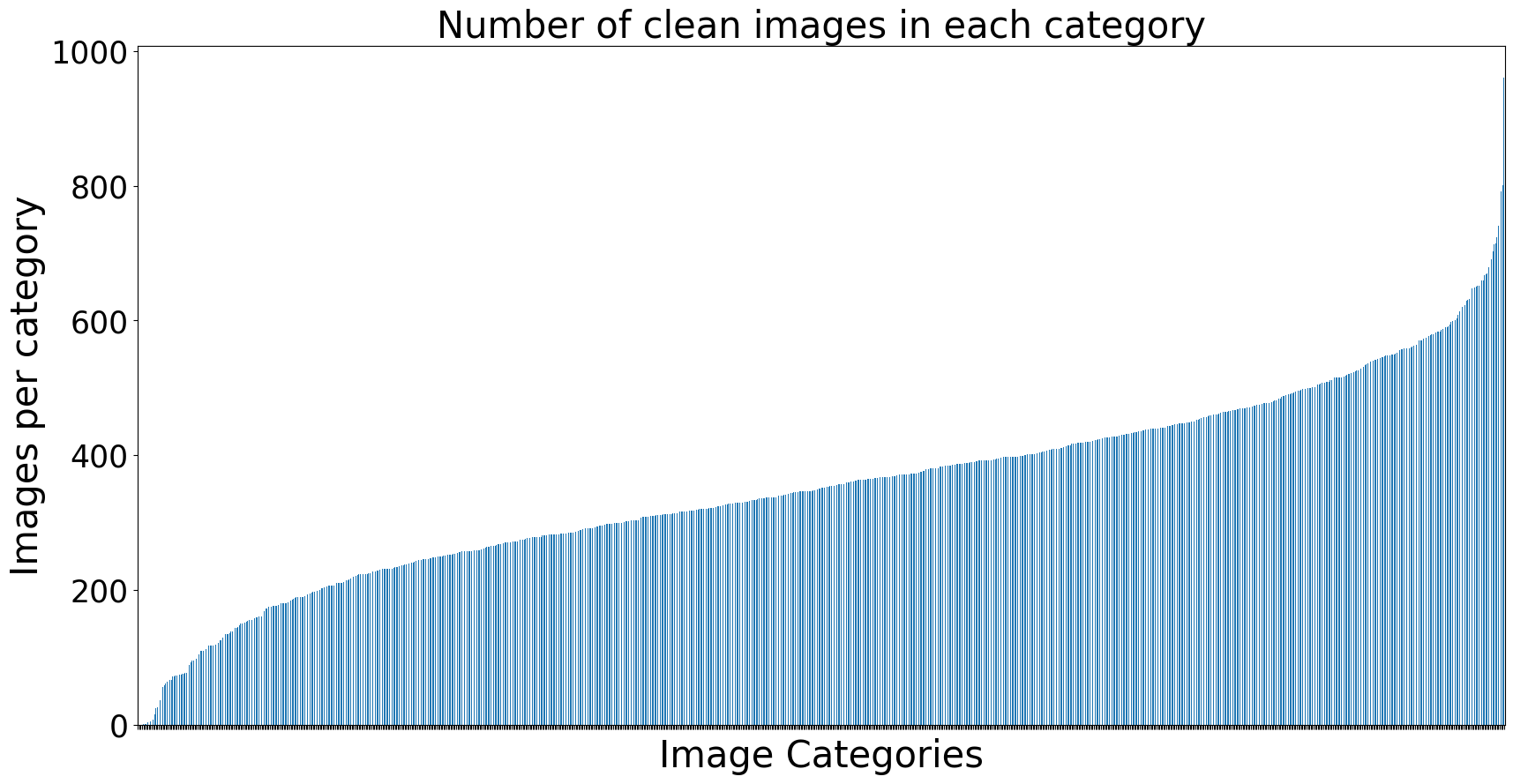}
  \caption{Histogram of the number of images per ImageNet category retrieved by Google search. There are natural imbalances in the data. Learning model under such imbalanced natural data distribution is an interesting problem to study.} 
  \label{fig:img_num}
  \vspace{-5mm}
\end{figure} 

\begin{figure*}[t]
\centering
\includegraphics[width=0.99\textwidth]{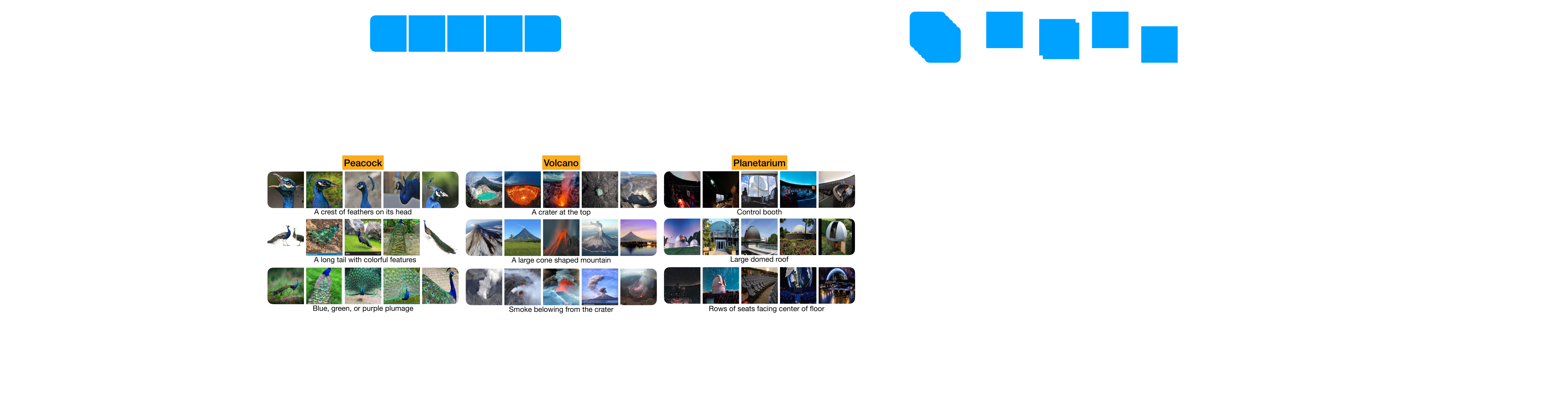}
  \caption{Examples for the images in our collected ImageNet$^+$ dataset, where each image corresponds to one category label and one rationale. We show three random categories, and for each category, we show three random rationales. The rationales are consistent with the image features that decide the object's category.} 
  \label{fig:google_example}
  \vspace{-5mm}
\end{figure*}

\subsection{Knowledge Transfer to Visual Domain}
Since not every rationale exists in every image of the given object, we cannot directly apply this list of rationales to images based on their category. We need  external knowledge to collect images containing specified rationales, so that we can correspond the language rationales to the visual rationals in images. Prior work~\cite{shen2022k} uses structured external knowledge from Wiktionary and WordNet to obtain images with text descriptions.  

In the era of the Internet, people often Google a new word if they don't know what the object --- the word refers to --- looks like. Once people see the image examples from Google, they quickly understand what the word refers to in the real world. Our method is motivated by this natural procedure, where our system Googles ``what the rationales look like'' to visually understand them.

We propose to leverage this external knowledge from web search engines, a larger-scale knowledge base than Wiktionary~\cite{shen2022k}. We use Google image search to obtain images that contain our specified visual features, and we find that the returned images match the language queries with high accuracy. The search engine serves as a cross-modal translator, translating the knowledge from the language domain to the visual domain by retrieving images.

We use the following queries to query images that belong to our specified category and contain our specified attribute:
\begin{lstlisting}[breakatwhitespace=true]
1. {category name} which has {attribute name}
2. {attribute name} of {category name}
3. a photo of {category name} because there is {attribute name}
\end{lstlisting}
and we retrieve the top 50 images returned by Google search.
More query sentences can be constructed if more images are needed, which we leave for future work. We remove the duplicated and incorrect images in preprocessing.

Google image search allows us to obtain images containing specific attributes. We perform the above automatic pipeline on ImageNet categories and obtain a dataset that contains images and the rationales that explain why this image is a particular category. For the dataset collected with our framework, we show a histogram of the number of images per ImageNet category in Figure~\ref{fig:img_num}. The data is S-curved due to the natural, non-interventional collection procedure. We also show examples of rationales and the retrieved images in Figure~\ref{fig:google_example}, where images can often be explained by rationales. Since our system is automatic, we in addition run our pipeline on the categories in CIFAR-10, CIFAR-100, Food101, Caltech101, and SUN datasets. Table~\ref{tab:imgnet_num} shows the scale of our collected images and compares them with prior datasets that are related to our work. Our pipeline allows richer rationales and is more extensive in the number of images and categories. To evaluate how accurate the Google image search can retrieve the images with the specified attribute as well as the category, we conduct a human study on the quality and consistency, which we describe in Section~\ref{sec:user_study}.

\begin{table*}[t]
  \caption{\label{tab:benchmark} Benchmarking six large-scale image-language pretrained models on doubly right recognition over six datasets. The accuracy for doubly right is gray-scaled. We \textbf{bold} the best accuracy. $\uparrow$ indicates higher number is better. For the same CLIP model, when model capacity increases from Res50 to H/14, the models' ability to get doubly right also increases, even though CLIP has never been trained on this metric. However, the doubly right accuracy on large-scale datasets is still low. For example, the double right accuracy is less than 1\% on ImageNet. While larger model provides higher accuracy for object classification, they often provide wrong rationales, as indicated by the increase in RW percentage. Our evaluation suggests that this new doubly right object recognition task is challenging for existing large-scale visual models.}
  \centering
  \scriptsize
  \begin{tabular}{l|llll|llll|llll}
    \toprule
    & \multicolumn{4}{c}{\scshape{CIFAR-10$^+$}} & \multicolumn{4}{c}{CIFAR-100$^+$} & \multicolumn{4}{c}{Food101$^+$} \\
    Model     & \cellcolor{Gray}RR $\uparrow$    & RW$\downarrow$   &WR$\downarrow$  & WW$\downarrow$ & \cellcolor{Gray}RR$\uparrow$     & RW $\downarrow$  &WR$\downarrow$  & WW$\downarrow$ & \cellcolor{Gray}RR$\uparrow$     & RW$\downarrow$   &WR$\downarrow$  & WW$\downarrow$ \\
    \midrule
    FLAVA & \cellcolor{Gray}29.44 & 53.04 & 5.83 & 11.68  &\cellcolor{Gray}3.57 & 59.72 & 5.44 & 31.26 &\cellcolor{Gray}3.95 & 55.78 & 4.86 & 35.42  \\
    CLIP-Res50 & \cellcolor{Gray}30.65 & 50.85 & 8.51 & 9.97  &\cellcolor{Gray}4.51 & 58.94 & 7.76 & 28.78 & \cellcolor{Gray}6.53 & 61.47 & 5.31 & 26.69 \\
    CLIP-Res101 & \cellcolor{Gray}30.41 & 50.61 & 8.76 & 10.21 & \cellcolor{Gray}5.09 & 60.31 & 7.41 & 27.19 & \cellcolor{Gray}5.23 & 64.65 & 4.61 & 25.46 \\
    CLIP-B/32 & \cellcolor{Gray}36.98 & 46.71 & 9.00 & 7.30 & \cellcolor{Gray}5.23 & 59.93 & 7.30 & 27.53 &5.48\cellcolor{Gray} & 63.57 & 5.41 & 25.53 \\
    CLIP-B/16 & \cellcolor{Gray}35.28 & 49.63 & 9.25  & 5.83 & \cellcolor{Gray}5.61 & 64.09 & 6.16 & 24.13 & 6.04\cellcolor{Gray} & 67.65 & 4.26 & 22.04 \\
    CLIP-L/14 & \cellcolor{Gray}\textbf{42.57} & 44.52 & 7.06 & 5.84 & \cellcolor{Gray}\textbf{6.43} & 63.71 & 7.73 & 22.13 & \textbf{5.73}\cellcolor{Gray} & 70.07 & 4.30 & 19.91 \\
    \midrule
     \toprule
    & \multicolumn{4}{c}{Caltech101$^+$} & \multicolumn{4}{c}{SUN$^+$} & \multicolumn{4}{c}{ImageNet$^+$} \\
    Model     & \cellcolor{Gray}RR  $\uparrow$   & RW  $\downarrow$ &WR $\downarrow$ & WW$\downarrow$ & \cellcolor{Gray}RR$\uparrow$     & RW $\downarrow$  &WR$\downarrow$  & WW$\downarrow$ & \cellcolor{Gray}RR$\uparrow$ & RW$\downarrow$   &WR $\downarrow$ & WW $\downarrow$\\
    \midrule
    FLAVA & \cellcolor{Gray}2.21 & 61.57 & 3.85 & 32.38 &  \cellcolor{Gray}\textbf{0.95} & 14.62 & 10.72 & 73.70 & \cellcolor{Gray}0.40 & 28.31 & 3.58 & 67.70   \\
    CLIP-Res50 & \cellcolor{Gray}4.13 & 61.60 & 5.31 & 26.69  &  \cellcolor{Gray}0.78 & 19.87 & 10.93 & 68.41 & \cellcolor{Gray}0.61 & 40.42 & 3.69 & 55.26 \\
    CLIP-Res101 & \cellcolor{Gray}4.57 & 64.24 & 4.61 & 25.46 & \cellcolor{Gray}0.89 & 21.29 & 11.68 & 66.75 & \cellcolor{Gray}0.65 & 42.74 & 3.64 & 52.98\\
    CLIP-B/32 & \cellcolor{Gray}4.51 & 65.92 & 5.42 & 24.15 & \cellcolor{Gray}0.86 & 23.32 & 10.72 & 65.16 & \cellcolor{Gray}0.68 & 42.69 & 3.87 & 52.76 \\
    CLIP-B/16 & \cellcolor{Gray}4.60 & 68.30 & 4.84 & 22.25 & \cellcolor{Gray}0.81 & 23.41 & 10.50 & 65.68 & \cellcolor{Gray}0.63 & 46.73 & 3.41 & 49.23 \\
    CLIP-L/14 & \cellcolor{Gray}\textbf{5.99} & 66.55 & 5.96 & 21.50 & \cellcolor{Gray}0.94 & 24.27 & 11.21 & 63.58 & \cellcolor{Gray}\textbf{0.72} & 50.00 & 4.34 & 44.94 \\
    \bottomrule
  \end{tabular}
\end{table*}

\begin{table*}[t]
  \caption{\label{tab:whyprompt_iid} {Gain of using why prompt to adapt models to perform doubly right object recognition. We evaluate the CLIP-H/14 model, except for ImageNet, where we use CLIP-B/32 due to the large dataset size. We use a deep prompt on ImageNet, which is indicated by *. We \textbf{bold} the best doubly right accuracy. Learning a why prompt significantly improves the models' ability to predict the right category as well as the right rationales, reducing the failures when the model predicts the right category with wrong rationales.}}
  \centering
  \scriptsize
  \begin{tabular}{l|l|ll|ll|ll|ll}
    \toprule

       &   & \multicolumn{2}{c}{\cellcolor{Gray}RR$\uparrow$  }   & \multicolumn{2}{c}{RW$\downarrow$ }   &\multicolumn{2}{c}{WR$\downarrow$ }  & \multicolumn{2}{c}{WW$\downarrow$ } \\
    Datasets & Prompt Length & \cellcolor{Gray}Baseline & \cellcolor{Gray}Ours      & Baseline & Ours &  Baseline & Ours  & Baseline & Ours  \\
    \midrule
    CIFAR-10$^+$ &3 & \cellcolor{Gray}42.57 & \cellcolor{Gray}\textbf{70.82} &  44.52 & 18.25 &  7.06 & 6.32 &  5.84 & 4.62 \\
    CIFAR-100$^+$ &3 & \cellcolor{Gray}6.43 & \cellcolor{Gray}\textbf{22.27} &  63.71 & 44.61 &  7.73 & 9.97 &   22.13 & 23.14 \\
    Food101$^+$ &3 & \cellcolor{Gray}5.73 & \cellcolor{Gray}\textbf{25.25} &  70.07 & 51.83 &  4.30 & 5.83 &  19.91 & 17.08 \\
    Caltech101$^+$ &3 & \cellcolor{Gray}5.99 & \cellcolor{Gray}\textbf{23.64} &  66.55 & 52.43 &  5.96 & 5.86 &  21.50 & 18.06 \\
    SUN$^+$ & 100 & \cellcolor{Gray}0.94 & \cellcolor{Gray}\textbf{6.70} &   24.27 & 8.29&  11.21 & 23.76 &  63.58 &61.24 \\
    ImageNet$^+$ & 30* & \cellcolor{Gray}0.68 & \cellcolor{Gray}\textbf{3.63} &  42.69 & 21.70 &  3.87 & 7.66 &  52.76 & 25.34 \\ 

    \bottomrule
  \end{tabular}
\end{table*}

\begin{table*}[t]
  \caption{\label{tab:zro-shot} {Zero-shot gain of using why prompt to adapt models to perform doubly right object recognition. For all datasets we use CLIP-H/14 model, except for ImageNet, where we use CLIP-B/32. We \textbf{bold} the best accuracy for doubly right object recognition. Our method can obtain better rationales on unseen datasets than baseline (CLIP) model.} }
  \centering
  \scriptsize
  \begin{tabular}{ll|ll|ll|ll|ll}
    \toprule
       \multicolumn{2}{c}{Zero-Shot Transfer}   & \multicolumn{2}{c}{\cellcolor{Gray}RR$\uparrow$ }     & \multicolumn{2}{c}{RW$\downarrow$}    &\multicolumn{2}{c}{WR$\downarrow$}   & \multicolumn{2}{c}{WW$\downarrow$}  \\
  Training Datasets &  Testing Datasets   & \cellcolor{Gray}Baseline & \cellcolor{Gray}Ours     & Baseline & Ours    & Baseline & Ours & Baseline & Ours \\
    \midrule
  CIFAR-100$^+$ &  CIFAR-10$^+$ & \cellcolor{Gray}42.47 & \cellcolor{Gray}\textbf{54.99}  & 44.52 & 33.58 & 7.06 & 4.87 & 5.84 & 6.57 \\
   CIFAR-100$^+$ & Food101$^+$ &  \cellcolor{Gray}5.73 & \cellcolor{Gray}\textbf{8.35} & 70.07 & 44.61 & 4.30 & 9.97 & 19.91 & 23.14 \\
   CIFAR-100$^+$ & Caltech101$^+$ &  \cellcolor{Gray}5.99 &  \cellcolor{Gray}\textbf{15.07}& 66.55 & 56.34 & 5.96 & 6.21 & 21.50 & 22.38 \\
    \midrule
   Caltech101$^+$ & CIFAR-10$^+$ & \cellcolor{Gray}42.47 &\cellcolor{Gray}\textbf{49.63} & 44.52 & 39.17 & 7.06 & 5.60 & 5.84 & 5.60 \\
    Caltech101$^+$ & CIFAR-100$^+$ & \cellcolor{Gray}6.43 & \cellcolor{Gray}\textbf{13.20} & 63.71 & 49.11 & 7.73 & 10.38 & 22.13 & 27.30 \\
   Caltech101$^+$ & Food101$^+$ &  \cellcolor{Gray}5.73 & \cellcolor{Gray}\textbf{7.36} & 70.07 & 61.05 & 4.30 & 4.49 & 19.91 & 26.69 \\
 \midrule
  SUN$^+$ &  CIFAR-10$^+$ & \cellcolor{Gray}42.47 & \cellcolor{Gray}\textbf{49.00} & 44.52 & 40.04 & 7.06 & 5.78 & 5.84 &5.18 \\
    SUN$^+$ & CIFAR-100$^+$ & \cellcolor{Gray}6.43 & \cellcolor{Gray}\textbf{13.11} & 63.71 & 49.63 & 7.73 & 8.32 & 22.13 & 28.93 \\
    SUN$^+$ & Food101$^+$ &  \cellcolor{Gray}5.73 & \cellcolor{Gray}\textbf{8.94} & 70.07 & 51.90 & 4.30 & 6.67 & 19.91 & 32.48 \\
    SUN$^+$ & Caltech101$^+$ &  \cellcolor{Gray}5.99 &  \cellcolor{Gray}\textbf{13.58} & 66.55 & 55.14 & 5.96 & 5.39 & 21.50 & 25.88  \\
 \midrule
  ImageNet$^+$ &   CIFAR-10$^+$ & \cellcolor{Gray}36.98 & \cellcolor{Gray}\textbf{38.68}&  46.71 & 43.80 & 9.00 & 9.25 & 7.30 & 8.27 \\
 ImageNet$^+$ & CIFAR-100$^+$ & \cellcolor{Gray}5.23 & \cellcolor{Gray}\textbf{15.67} & 59.93 & 39.89 & 7.30 & 9.51 & 27.53 & 55.57 \\
 ImageNet$^+$ & Food101$^+$ &\cellcolor{Gray}5.48 &\cellcolor{Gray}\textbf{8.31} & 63.57 & 46.59 & 5.41 & 5.97 & 25.53 & 39.12 \\
 ImageNet$^+$ & Caltech101$^+$ & \cellcolor{Gray}4.51 &\cellcolor{Gray}\textbf{16.71} & 65.92 & 45.42 & 5.42 & 7.66 & 24.15 & 30.20 \\
 ImageNet$^+$ & SUN$^+$ & \cellcolor{Gray}0.86 &\cellcolor{Gray}\textbf{1.98} &  23.32 & 7.02 & 10.72 & 14.96 & 65.16 & 76.90  \\
\bottomrule
  \end{tabular}
\end{table*}

\subsection{Learning the Why Prompt}

Using external knowledge from both the language model and Google, we show we can collect image datasets containing both the category and rationales. Our pipeline allows us to evaluate the quality of this doubly right task on a large scale over several datasets for the first time. Our results show that doubly right object recognition is a challenging task for large visual models, such as CLIP.

We then seek a way to adapt and improve large pre-trained models so that they can provide the correct rationale for the predictions. 
Motivated by the prompting in natural language processing, we propose constructing a visual ``why" prompt that instructs the model to produce the right rationales for the prediction. We optimize the visual prompt to ask ``what are the visual rationales that explains the prediction of this image''. We use the following input prompts or deep prompts to adapt the model.


\noindent\textbf{Prompt Design.} We study how to adapt transformer-based models since they are state-of-the-art. Our visual \emph{why} prompts append additional tokens $P_k$ to the input sequence of the vision transformer:
\begin{equation}
        \x = [\x; P_0, P_1, ..., P_k]
\end{equation}
The remaining transformer parameters and computations are kept the same as the original.

\noindent\textbf{Deep prompt.} Besides adding context to the input sequence, the deep prompting method~\cite{jia2022visual} adds prompts to the intermediate layers in the transformer, which is a more powerful adaptation method than single-layer  input prompt. Let $\x_i$ be the latent token sequence of the $i-$th layer in the transformer. We can add a prompt to each latent layer:
\begin{align}
        &\x_i = [\x_i^0; P_i, P_i^1, ..., P_i^k] \\
        &\x_{i+1} = \text{Head}(\x_i) 
\end{align}
where the Head indicates the transformer block. We find deep prompt is particularly useful for learning on large-scale datasets, such as ImageNet.

\noindent\textbf{Learning Objective.} 
We now introduce the training objective to instruct models for doubly right object recognition. We use cross-modal image-to-text contrastive learning to train the model for the right rationales. Using our above pipeline, we have collected a set of images and their rationales, which we denote as $\{(\x_i, \t_i)\}$. In contrast to the prior text prompts for object recognition, which uses:
\begin{lstlisting}[breakatwhitespace=true]
This is a photo of [CATEGORY]
\end{lstlisting}
we instead create the training text prompt to be:
\begin{lstlisting}[breakatwhitespace=true]
This is a photo of [CATEGORY] because there is [RATIONALE]
\end{lstlisting}
This allows the model to learn what is the correct rationales that explain the category prediction. 

We use the pretrained image-language model, where we encode the image with an image encoder $F_\theta$ and the corresponding rationale with a text encoder $T$. To train the model to rank the correct rationales higher than the other negative rationales, we minimize the following image-to-text contrastive loss function,
\begin{equation}
    \mathcal{L}_s(\x, \t, \y) = -\mathbb{E}_{i,j}\left[\y_{ij}
    \log \frac{\exp(\mathrm{cos}(\z_i^{(I)}, \z_j^{(T)})/\tau)}{\sum_{k}\exp(\mathrm{cos}(\z_i^{(I)}, \z_k^{(T)})/\tau)}  
    \right], \label{eq:multimodalcontrastive}
\end{equation}
where  $\z_i^{(I)} = F_\theta(\x_i)$ and $\z_i^{(T)} = T(\t_i)$ are the features from the image and text, respectively. $\y_{ij}$ indicates which image-text are paired in the dataset and which are not. We set $\y_{ij} = 1$ when the image and text are from the same data point. $\mathrm{cos}$ denotes the cosine similarity function. $\tau$ is the temperature hyperparameter to scale the confidence of the prediction.
We then use gradient descent to optimize the visual prompt such that this loss is minimal.

\subsection{Evaluation Metric}
We define the following metric to evaluate the models' predictions as well as their rationales. We formulate producing the rationales as a ranking task, where we provide the model sentence descriptions containing a pair-wise combination of all the categories and rationales, and the image-language model will retrieve the closer ones. Specifically, we present the model with a sentence in this format:
\begin{lstlisting}[breakatwhitespace=true]
This is a photo of [CATEGORY] because there is [RATIONALE]
\end{lstlisting}
and ask the model to return the sentence that has the closest representation to the visual embeddings.
Since one image may have multiple rationales to explain the category, following the standard practice~\cite{grauman2022ego4d}, we use a top-K accuracy --- if the top K rationales include the ground truth, then it is counted as correct. The predicted category is based on the majority vote of the top-K reterieved categories~\cite{menon2022visual}. We denote the metric as follows:
\begin{enumerate}[label=(\arabic*),leftmargin=2em, itemsep=0mm]
    \item Right classification with right rationale (RR);
    \item Right classification with wrong rationale (RW); 
    \item Wrong classification with right rationale (WR);
    \item Wrong classification with wrong rationale (WW)
\end{enumerate}
We desire a high accuracy for RR and a low percentage for RW, WR, and WW.
We will evaluate the above metric on our collected dataset where rationale ground truth is provided. We select 20\% of the data as the hold-out test set. We will train the model on the remaining 80\% data.

%% file: src/Experiments.tex
\begin{figure*}[t]
\centering
\vspace{-5mm}
\includegraphics[width=0.99\textwidth]{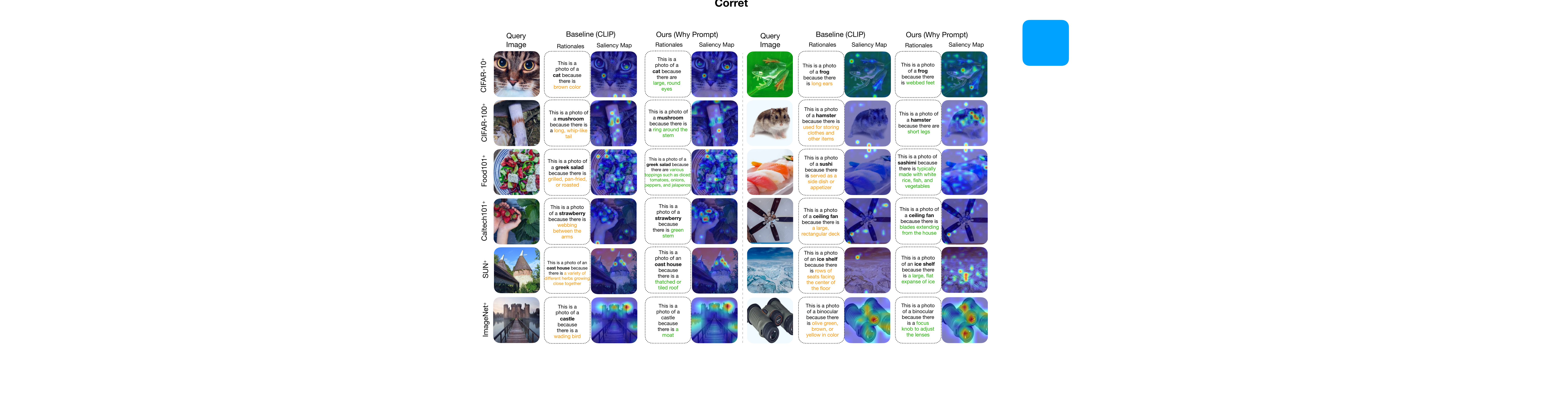}
\vspace{-3mm}
  \caption{Visualization for the doubly right recognition task. For each row, we show image examples from categories of one dataset. In columns 2, 4, 7, and 9, we show the rationales produced by the model to explain the prediction. In column 3, 5, 8, and 10, we show the saliency map~\cite{chefer2021generic} that models look to produce the prediction and rationales. While the state-of-the-art H/14 CLIP model produces the correct category with the wrong rationales, our method can produce the correct category with the right rationales. In addition, our visual prompt also adapts the model to use the right image region to produce the prediction.} 
  \label{fig:viz_ours}
\end{figure*} 

\begin{figure*}[t]
\centering
\vspace{-3mm}
\includegraphics[width=0.99\textwidth]{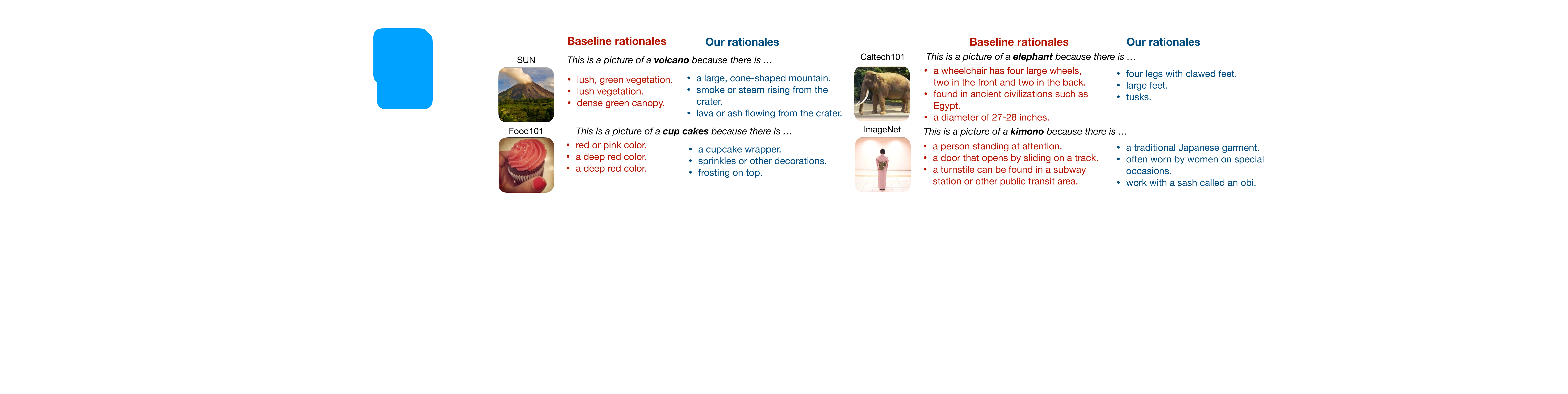}
\vspace{-5mm}
  \caption{Visualization for top 3 rationales. We show the original examples from SUN, Caltech101, Food101, and ImageNet datasets with high-resolution images. Even though we do not annotate rationales for images in these datasets, our method can successfully transfer the rationales learned from our data and apply them.} 
  \label{fig:word}
  \vspace{-5mm}
\end{figure*} 

\section{Experiment}

We evaluate six large-scale image-language pretrained models on our doubly right benchmark. We find that the model often produces the wrong rationales for the predictions. We then show that our \emph{why} prompts significantly improves the models' ability to produce the right rationales quantitatively and qualitatively. Lastly, we show that we can have a hierarchy of visual rationales where the model can provide sub-rationales. 

\subsection{Benchmark Existing Models}


We start our investigation by evaluating existing large-scale image language models~\cite{radford2021learning, singh2022flava} on our collected doubly right dataset. Our evaluation includes images in the category of CIFAR-10, CIFAR-100, Food101, Caltech101, SUN, and ImageNet. Since we recollect the image based on the category name of those dataset, we use $^+$ to denote our collected dataset. We set $K=5$ for the top-K accuracy of the doubly right prediction. We study FLAVA~\cite{singh2022flava} and five variants of CLIP~\cite{radford2021learning}, where evaluation results are in Table~\ref{tab:benchmark}. We find that for the same CLIP model, increasing their capacity from Resnet 50~\cite{resnet_2015} to ViT Huge/14~\cite{vit} generally improves performance in retrieving the right rationales, even if the model has never been trained on the doubly right task. FLAVA model performs worse than all CLIP model variants, except on SUN$^+$. Despite the high classification accuracy, doubly right object recognition is challenging for all models, where models produce incorrect rationales more often than the correct ones.
Specifically, the best CLIP-H/14 model only obtains 1\% accuracy on ImageNet$^+$ doubly right recognition task. Our evaluation shows that doubly right object recognition is still an open challenge for the large-scale dataset.





\subsection{Why Prompt for Visual Rationales}
To adapt the vision model so that they can perform doubly right object recognition, we apply our ``why'' prompting to visual foundation models. On ImageNet$^+$, we adapt the CLIP-B/32 due to the large size of the dataset. We use deep prompt with prompt length 30 for each of the 12 layers. We train 10 epochs with a learning rate of 10. For all the other datasets, we adapt the best CLIP-H/14 model. We train model for 100 epochs with a learning rate of 40. We use a prompt size of 3, except for SUN where we use 100. For each dataset, we train on 80\% of the data and test on the hold out 20\% test data.

We show our results in Table~\ref{tab:whyprompt_iid}. By finding a why prompt to adapt the visual model, we significantly improve the models' ability to produce the right predictions as well as the right rationales, up to \textbf{28} points. Using our prompt, the model also sometimes predicts the wrong categories when the rationales are wrong (higher WW), suggesting our \emph{why} prompt achieve higher consistency in reasoning between category and rationales. Our prompt method is still effective when the prompt size is as small as 3, containing minimal parameters, which is lightweight to apply. In addition, learning our why prompt can maintain the classification accuracy on the original ImageNet validation set, where CLIP-B/32 obtains an accuracy of 59.16\%, and we obtain an accuracy of 59.20\%.

\noindent\textbf{Zero-shot Transferability for Why Prompt.} 
We find that the above generated ``why'' also transfers to unseen datasets and categories. In Table~\ref{tab:zro-shot}, we show the doubly right accuracy obtained by zero-shot transfer, where we obtain up to 12 points gain. For example, by learning the model on the SUN$^+$ dataset which contains natural scenes, our method can teach CLIP to provide the correct rationales for food images (Food101$^+$), by 3 points better. This experiment shows our \emph{why} prompt can adapt large scale models to produce correct predictions with right rationales, and generalize to novel categories, suggesting the effectiveness of our method.



\noindent\textbf{Visualizations on Doubly Right dataset.}
We visualize images from the test set of our benchmark. In Figure~\ref{fig:viz_ours}, we show the top 1 rationales retrieved by the baseline and our prompted model. Our method often produces a correct explanation, while the baseline does not. We also visualize the saliency map corresponding to the rationales~\cite{chefer2021generic}, where we can also see that the why prompt instructs the model to look at the right region to explain the prediction.

\noindent\textbf{Qualitative results on the original dataset.} In addition to evaluating the doubly right performance on our collected dataset above, we also evaluate the established, original data. We study ImageNet, SUN, Caltech, and Food101 datasets since they contain high-resolution images that allow detailed explanations. Though we cannot directly evaluate the doubly right accuracy of those dataset images since they do not contain annotated ground truth, we show visualizations for the rationales generated on those datasets. In Figure~\ref{fig:word}, we find our learned ``why'' prompt also transfers well in improving the doubly right performance on images in those datasets.

\subsection{Analyzing Hierarchical Rationales}
The rationales provided by our approach may not be the basic explanations. For example, with the above method, we can explain that this is a photo of a dog because there are four legs. We are curious whether the model can also provide another level of visual reasoning for why the four legs cause the image to be a dog.
We explore whether our framework can provide even more fundamental explanations for the generated rationales. 

\begin{figure}[t]
\centering
\includegraphics[width=0.45\textwidth]{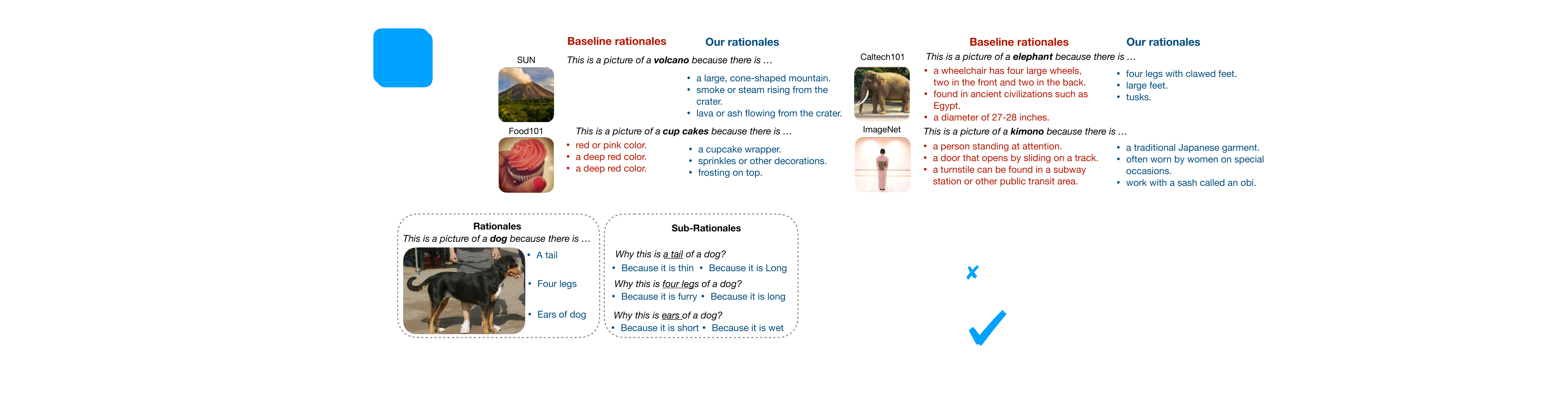}
\vspace{-3mm}
  \caption{Visualizations for the hierarchical rationales on CIFAR-10$^+$. Our method obtains reasonable sub-rationales by learning an additional sub \emph{why} prompt.} 
  \label{fig:h2}
  \vspace{-5mm}
\end{figure} 

To obtain sub-rationales, we query GPT3 with:
\begin{lstlisting}[breakatwhitespace=true]
Q: What are useful visual features for distinguishing a {attribute name} of a {category name} in a photo?
A: There are several useful visual features to tell there is a {attribute name} of a {category name}  in a photo:
\end{lstlisting}

After getting the sub-rationales through language chain of thought, we transfer the knowledge to visual domain by Google query:
\begin{lstlisting}[breakatwhitespace=true]
A photo of {CATEGORY}, because there is {sub-level attribute name} {attribute name}
\end{lstlisting}
where we collect 10583 images that contain attributes of specific patterns denoted by the sub-rationales. We split the dataset into an 80\% training set and 20\% of testing set.

We train our why prompt with the same objective in Equation~\ref{eq:multimodalcontrastive}. We then evaluate the correctness of the rationales and the sub-rationales. We experiment on CIFAR-10$^+$ and train the model for 25 epochs.  When categories, rationales, and sub-rationales are correct, it is counted as doubly right (RR). In Table~\ref{tab:hierarchy}, baseline CLIP has 0 accuracy. Our method improves RR by 26 points. In Figure~\ref{fig:h2}, we visualize this hierarchical rationale, where our method can produce sub-rationales that explains the rationales hierarchically.

\begin{table}[t]
  \caption{\label{tab:hierarchy} Accuracy for hierarchical visual rationales. On CIFAR-10$^+$ categories, we evaluate CLIP and our models' ability to get the category, the rationales, and the rationales' explanations correct (RR). Our method produces better sub-rationales than CLIP.}
  \centering
  \small
  \begin{tabular}{lllll}
    \toprule
    & \cellcolor{Gray}RR     & RW   &WR  & WW \\
    \midrule
    CLIP-H/14 & \cellcolor{Gray}0.05 & 38.14 & 0.20 & 61.61 \\
    Ours & \cellcolor{Gray}\textbf{26.19} & 53.86 & 1.58 & 18.35 \\
    \bottomrule
  \end{tabular}
  \vspace{-5mm}
\end{table}

\subsection{User Study}\label{sec:user_study}
To evaluate the quality of our doubly right dataset, we conduct a human study. We show 200 images randomly sampled from our dataset and ask the user to check whether the category and rationales match what's inside the image. We conduct this study on 20 participants. Since the task requires visual reasoning, it takes 20 to 40 minutes for one participant to complete the study. On average, the user thinks  86.2\% of the annotation is correct, with a standard deviation of 8.5. This shows our pipeline can collect visual data with visual rationales of reasonably high quality.








%% file: src/conclusion.tex
\section{Conclusion}
In this paper, We study an essential yet under-explored visual problem, getting the correct rationales for the predictions, which we name as the ``doubly right'' object recognition task. We construct large-scale datasets containing categories from various datasets with rich rationales, which allows us to evaluate this doubly right metric directly. Our work proposes a pipeline that transfers the rationales knowledge from language models to visual models, which significantly improves the doubly right object recognition accuracy on both seen and unseen categories. Our work provides a benchmark and algorithm that allows the visual recognition field to push this doubly right task forward.

\section*{Acknowledgement}
This research is based on work partially supported by the DARPA GAILA program, the DARPA KAIROS program, the NSF NRI Award \#2132519, a GE/DARPA grant, a CAIT grant, and gifts from JP Morgan, DiDi, and Accenture. We thank Jianan Yao on feedback for user study. 
